\documentclass[10pt,twocolumn,letterpaper]{article}

\usepackage[pagenumbers]{cvpr} 

\definecolor{cvprblue}{rgb}{0.21,0.49,0.74}
\usepackage[pagebackref,breaklinks,colorlinks,allcolors=cvprblue]{hyperref}
\usepackage{footmisc}
\usepackage[table]{xcolor} 
\usepackage{multirow}
\usepackage{mathtools}

\newcommand{\mymethod}{FaithTrace\xspace}

\usepackage[most]{tcolorbox}
\tcbuselibrary{listings,breakable,skins}

\newtcblisting{promptverb}{
  enhanced,
  breakable,
  listing only,
  colback=black!2,
  colframe=black!40,
  boxrule=0.4pt,
  arc=2pt,
  left=1.2mm,right=1.2mm,top=1.0mm,bottom=1.0mm,
  fonttitle=\bfseries,
}

\title{Zero-Shot Faithful Textual Explanations via Directional-Derivative Influence on Predictions}

\author{
Toshinori Yamauchi$^{1}$ \and
Hiroshi Kera$^{1,2}$ \and
Kazuhiko Kawamoto$^{1}$ \and
$^{1}$Chiba University \quad $^{2}$National Institute of Informatics\\
{\tt\small t.yamauchi@chiba-u.jp, kera@chiba-u.jp, kawa@faculty.chiba-u.jp}
}

\begin{document}
\maketitle

\begin{abstract}
Zero-shot textual explanations aim to make image classifiers more transparent by probing their internal representations, without relying on task-specific supervision or LVLMs.
However, existing methods often miss the features that truly drive the prediction,
resulting in limited \textit{faithfulness} to the evidence underlying the model's decision.
To address this, we propose FaithTrace.
Motivated by the idea that faithful explanations should describe concepts that strongly influence the prediction, 
FaithTrace directly measures how much the representation induced by the explanation changes the class logit.
We introduce an influence score, computed as the directional derivative of the class logit along the text-induced direction in the classifier's feature space, 
and use it as a proxy for faithfulness.
Moreover, we extend this influence score into quantitative evaluation metrics, 
helping fill the gap in faithfulness evaluation for textual explanations.
Experiments show that FaithTrace yields more faithful explanations than baselines, 
facilitating a more accurate understanding of the model.
The code will be publicly released.
\end{abstract}

\section{Introduction}
\label{sec:intro}

Textual explanations describe why an image classifier makes a particular prediction,
which are important for building reliable vision systems.
Zero-shot approaches~\cite{text_to_concept,zsnle,texter} are a promising way to generate such explanations for off-the-shelf classifiers.
They operate directly on the classifier's existing representations and require no additional training or task-specific annotations.
This contrasts with supervised approaches, such as concept-bottleneck models~\cite{cbn, label_free_cbn, hybrid_cbn} and natural language explanation models~\cite{nlx_gpt, uni_nlx}, as well as large vision--language models (LVLMs)~\cite{pmlr-v162-li22n, NEURIPS2023_6dcf277e, openai2024gpt4technicalreport}.
Zero-shot approaches help users understand which cues the model relies on and whether those cues are semantically meaningful.

\begin{figure*}[t]
    \centering
      \includegraphics[width=\linewidth]{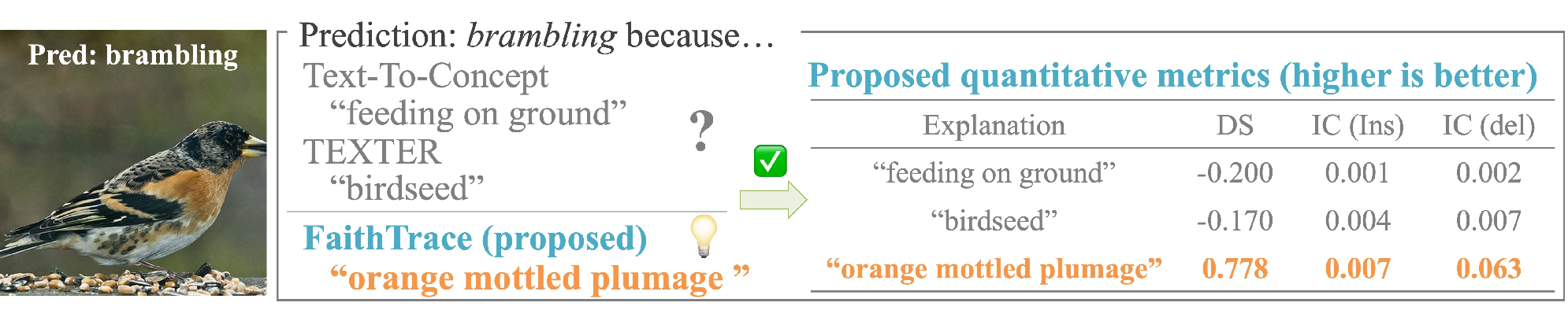}
      \caption{Comparison of explanations produced by Text-To-Concept~\cite{text_to_concept}, TEXTER~\cite{texter}, and the proposed \mymethod.Existing methods tend to yield generic features, whereas \mymethod highlights distinctive, class-specific evidence. 
      The right table reports our evaluation metrics—directional score (DS) and the summed insertion/deletion influence curves (IC (ins) and IC (del), in units of $10^{-1}$; see \cref{sec:evaluation_metric})—showing that the explanation from \mymethod achieves the highest scores.}
  \label{fig:intro}
\end{figure*}

\par
However, the \textit{faithfulness} of existing zero-shot explanations often remains limited or unclear, which can mislead users about the model's behavior.
A key reason is that these methods rely on external VLM-space similarity (e.g., CLIP) to provide explanations, 
without measuring their effect on the classifier's prediction.
Such similarity is correlational: it captures semantic match to the image but not whether the described cues drive the prediction.
For example, Text-To-Concept~\cite{text_to_concept} aligns global image features with CLIP~\cite{clip} and often outputs generic, non-decision-driving descriptions.
TEXTER~\cite{texter} introduces a concept image to better isolate decision-critical features, 
yet it still relies on CLIP-space similarity, making faithfulness hard to guarantee.
Indeed, in \cref{fig:intro}, for \textit{brambling}, both methods focus on the seeds in the lower part of the image and produce explanations such as ``feeding on ground'' and ``birdseed.''
These descriptions are generic for many birds and do not explain why the classifier predicts \textit{brambling}.

\par
To address this limitation, we propose \textbf{FaithTrace}
(zero-shot \textbf{FAITH}ful \textbf{T}extual explanations via di\textbf{R}ectional-deriv\textbf{A}tive influen\textbf{CE} on predictions).
The proposed method is motivated by the idea that faithful explanations should describe concepts that strongly influence the prediction.
Crucially, we use a VLM to connect text and vision, but validate explanations by directly measuring their influence on the classifier's prediction in its feature space.
To this end,
we introduce an influence score,
computed via the directional derivative of the class logit along the text-induced direction in the classifier's feature space, 
which serves as a proxy for the faithfulness in the explanations.
We compose explanations from texts with high influence scores, thereby highlighting decision-driving factors.
As shown in \cref{fig:intro}, \mymethod produces ``orange mottled plumage,'' a distinctive attribute of \textit{brambling}.
Moreover, we extend the influence score into quantitative faithfulness metrics, complementing existing protocols for textual explanations of image classifiers.
These metrics enable objective evaluation beyond human intuition.
In \cref{fig:intro}, the proposed metrics---directional score (DS) and summed insertion/deletion influence curves (IC (ins) and IC (del); see \cref{sec:evaluation_metric})---quantitatively support the effectiveness of \mymethod.

\par
In experiments, 
we evaluate the proposed method on a diverse set of CNN- and Transformer-based classifiers.
Across quantitative metrics, our approach consistently and clearly outperforms prior zero-shot baselines.
Qualitatively, it produces more faithful explanations that better reflect the model's decision basis, enabling more accurate interpretation of classifier behavior.
Our contributions are summarized as follows.
\begin{itemize}
  \item 
  We propose \mymethod,
  a zero-shot framework for faithful textual explanations that plugs into diverse off-the-shelf classifiers by directly measuring an explanation's influence on the prediction in the classifier's internal space.

  \item 
    We propose an evaluation metric based on the influence score that quantitatively assesses explanations from the perspective of faithfulness, 
    complementing existing protocols that remain limited in this field.

  \item 
    Extensive experiments on various CNN and Transformer classifiers show that \mymethod consistently surpasses prior zero-shot baselines.
    While existing methods often produce generic descriptions shared across many classes, \mymethod yields distinctive explanations that better reflect the decision-driving evidence, helping diagnose spurious biases in the model.
\end{itemize}

\section{Related work}
\label{sec:related_work}

Textual explanation methods for image classification models have been explored along several lines. 
One broad stream comprises concept bottleneck models~\cite{8099837,tcav,cbn,label_free_cbn,labo},
natural language explanation models~\cite{generate_visual_explanation,park_nle,nlx_gpt,uni_nlx},
and LVLM-based reasoning systems~\cite{pmlr-v162-li22n,NEURIPS2023_6dcf277e,openai2024gpt4technicalreport}, 
which require concept supervision, model retraining, or training dedicated explanation generators on annotated rationales.
We provide a more detailed review of these approaches in Appendix~\ref{sec::more_releted_works}.

\par
In this work, by contrast, 
we focus on zero-shot explanations for image classifiers trained only on images and class labels, without retraining, additional annotations, or external explanation models.
In the remainder of this section, we first review such zero-shot methods and discuss their limitations (\cref{sec:related_works_zero_shot}),
and then describe existing quantitative evaluation metrics in this field and their limitations (\cref{sec:related_works_evaluation}).

\subsection{Zero-shot textual explanations}
\label{sec:related_works_zero_shot}

Zero-shot textual explanations aim to explain pretrained models without additional supervision or retraining.
While several approaches rely on vision--language models~\cite{menon,shtedritski,zs-a2t,interpreting_clip}, 
methods tailored to image classifiers remain relatively scarce.
A prominent line aligns a classifier's feature space with CLIP~\cite{text_to_concept,beyond_clip,texter}: 
Text-To-Concept~\cite{text_to_concept} introduces a linear aligner between the classifier's feature space and CLIP, enabling direct comparison between visual features and text embeddings, thereby producing explanations that describe visible content.
ZSNLE~\cite{zsnle} instead maps text embeddings into the classifier space,
and another work~\cite{beyond_clip} further develops this CLIP-alignment framework by decomposing representations into component-wise contributions aligned to textual attributes.
However, these methods rely on global features rather than decision-critical evidence, which can yield explanations that are not faithful to the classifier's decision process, and they do not explicitly target faithfulness.
Closest to our goal, 
TEXTER~\cite{texter} seeks to improve faithfulness by translating an isolated, 
feature-visualized concept image~\cite{fv,maco,vital} into text; 
still, such indirect representations based on feature visualizations may allow elements that are not directly involved in the decision, 
so the resulting text can deviate from the classifier's true decision basis.
In contrast, our method directly measures influence on the prediction within the classifier's internal representations, producing explanations that better match the model's true decision basis.

\subsection{Evaluation of textual explanations}
\label{sec:related_works_evaluation}

Quantitative metrics for the faithfulness of textual explanations remain limited.
Prior work scores explanations by semantic similarity to the input image~\cite{zsnle,texter}, e.g., cosine similarity between CLIP image and text embeddings.
Such metrics mainly reward describing visible content and do not test whether the explanation reflects evidence that drives the classifier's prediction.
Another line uses human-annotated masks for regions referenced by an explanation and measures the resulting change in the prediction score~\cite{NLE_faithful}.
However, region masking cannot separate concept-specific evidence (e.g., ``snake body'' vs.\ ``scaly texture'') because both lead to nearly identical masked pixels.
It is also biased toward maskable cues: relational concepts (i.e., cues defined by relationships between visual features rather than a single localized concept) are hard to localize and thus fall outside the scope of this protocol.
As a result, this evaluation is confounded by maskability, making faithfulness harder to assess.
To address this, we propose influence-based metrics that directly quantify the effect of text-induced feature directions on the class logit in the classifier's feature space.

\section{Method}
\label{sec:method}
We first introduce the problem setup to clarify the goal of our approach and the associated challenges (\cref{sec:setup}).
We then describe the proposed method (\cref{sec:proposed_method}).
Finally, as an additional contribution, we describe influence-based quantitative metrics for evaluating the faithfulness of textual explanations (\cref{sec:evaluation_metric}).

\subsection{Problem setup}
\label{sec:setup}
We consider a fixed $C$-class image classifier $g\circ f:\mathcal{X}\to\mathbb{R}^{C}$.
The vision encoder $f:\mathcal{X}\to Z_f$ maps an input image $x\in\mathcal{X}$ to a feature vector $z_f=f(x)\in Z_f$, and the classification head $g:Z_f\to\mathbb{R}^{C}$ outputs logits $y=g(z_f)$.
We denote the logit for class $c$ by $y_c=g_c(z_f)$.
The classifier is trained only on images and is not coupled with any language model.
Given such a classifier $g\circ f$, 
our goal is to provide an explanation of its decision (prediction) in natural language.

\par
In this study, 
we regard a textual explanation as highly faithful if the feature direction it describes strongly affects the prediction.
Formally, 
for the logit of class $c$, $g_c(z_f)$, 
we define the influence score along a direction $\hat{v}_t(x)\in Z_f$ as
\begin{align}
I_c(x; \hat{v}_t(x)) = g_c\bigl(z_f+ \epsilon \hat{v}_t(x)\bigr)-g_c\bigl(z_f\bigr),
\label{eq:faithfulness}
\end{align}
where $\epsilon>0$ is a small step size and $\hat{v}_t(x)$ is a normalized feature-space direction vector in $Z_f$ induced by the explanation $t$.
This score measures how much the class-$c$ logit changes when the representation is perturbed along $\hat{v}_t(x)$; 
explanations with large positive $I_c(x; \hat{v}_t(x))$ are thus regarded as highly faithful.
As detailed later in \cref{sec:proposed_method}, 
this influence score can be expressed in terms of the directional derivative,
and our goal is to produce explanations with large positive scores.

\par
To achieve this goal, we face two main challenges.
The first is to relate the latent feature space of the target classifier to a language domain.
For this, we use a VLM with an image encoder $E_{\text{img}}: \mathcal{X} \to Z_{E_{\text{img}}}$ and a text encoder $E_{\text{text}}: \mathcal{T} \to Z_{E_{\text{text}}}$, and align $Z_f$ with $Z_{E_{\text{img}}}$ so that classifier features can be associated with text via the VLM.
The second challenge is, within the classifier's feature space, to obtain a direction vector $\hat{v}_t(x)$ induced by a textual explanation and to quantitatively evaluate its influence $I_c(x; \hat{v}_t(x))$ on the prediction.
In \cref{sec:proposed_method}, we describe our method focusing on these two aspects.

\begin{figure*}[t]
    \centering
    \includegraphics[width=\linewidth]{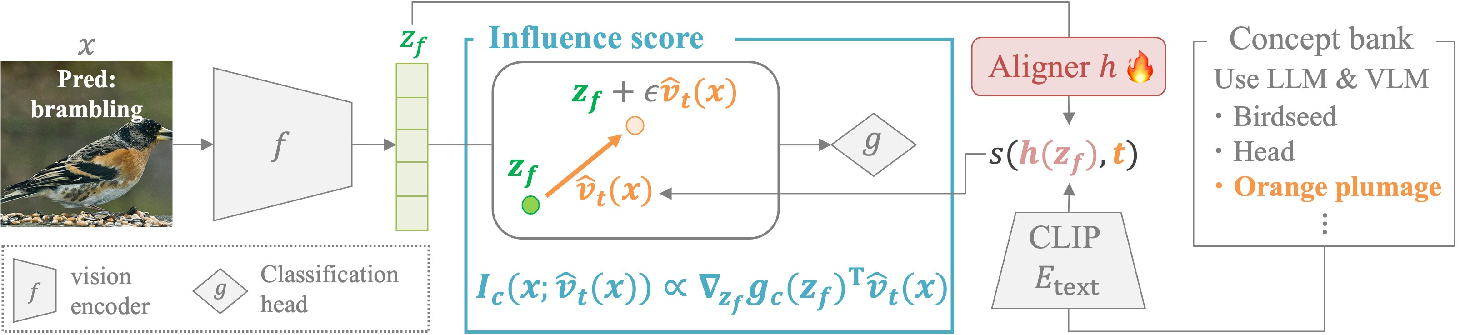}
    \caption{Overview of \mymethod. Given an input image $x$,
    we align the classifier feature $z_f$ to the CLIP embedding space via a learned affine aligner $h$.
    For a candidate text $t$ from a concept bank $\mathcal{B}(x,c)$,
    we obtain a text-induced direction vector $\hat{v}_t(x)$ by differentiating the CLIP similarity $s(h(z_f),t)$ with respect to $z_f$. 
    We then compute an influence score for class $c$ using the directional derivative $\nabla_{z_f} g_c(z_f)^{\top} \hat{v}_t(x)$, 
    and output the top-$k$ scoring texts as faithful explanations.
    The aligner $h$ is trained without extra annotations and used only for explanation, leaving inference and accuracy unchanged.}
    \label{fig:mymethod}
\end{figure*}

\subsection{\mymethod}
\label{sec:proposed_method}
Figure~\ref{fig:mymethod} gives an overview of the proposed method.
First, 
we describe how to align the latent feature space of the target classifier with a language domain.
We then describe how to compute the direction vector $\hat{v}_t(x)$ and the influence score $I_c(x;\hat{v}_t(x))$ in \cref{eq:faithfulness}, and how we use them to provide textual explanations.

\subsubsection{Vision-language space alignment}
\label{sec:alignment}
We adopt CLIP~\cite{clip} as the VLM and learn to map the classifier feature space onto the CLIP vision feature space.
Following Text-To-Concept~\cite{text_to_concept}, 
we introduce an affine aligner $h(z_f) = Wz_f + b$ and train it by minimizing
\begin{align}
\min_{W,b}
\frac{1}{|\mathcal{D}_{\text{train}}|}
\sum_{x \in \mathcal{D}_{\text{train}}}
\left\| W z_f+b-E_{\text{img}}(x) \right\|_2^2,
\label{eq:train_h}
\end{align}
where $\mathcal{D}_{\text{train}}$ is the training dataset, $W,b$ are the learnable parameters, 
and $z_f = f(x)$ denotes the classifier feature for input image $x$.

\par
After alignment, 
the mapped classifier feature $h(z_f)$ lies in the CLIP image-embedding space, 
where text embeddings are also defined.
As a result, 
an input image $x$ and an arbitrary text description $t$ can be compared in this joint space via cosine similarity:
\begin{align}
s(h(z_f),t)=\hat{z}_h^{\top}\hat{z}_{\text{text}},
\label{eq:sim_score_def_short}
\end{align}
where $\hat{z}_h = \frac{h(z_f)}{\|h(z_f)\|_2}$ and $\hat{z}_{\text{text}} = \frac{E_{\text{text}}(t)}{\|E_{\text{text}}(t)\|_2}$.
In the next section, 
this similarity is further exploited to derive a text-induced direction $\hat{v}_t(x)$ for computing the influence score $I_c(x; \hat{v}_t(x))$.

\subsubsection{Computation of the influence score}
\label{sec:influence_score}
As discussed in \cref{sec:setup},
our goal is to generate textual explanations that capture evidence strongly influencing the prediction score.
To this end, we evaluate the influence score $I_c(x; \hat{v}_t(x))$ defined in \cref{eq:faithfulness},
which requires associating each textual description $t$ with a direction vector in the classifier's feature space.
We define this direction as the gradient direction that increases the similarity $s(h(z_f),t)$ between visual and textual features.
Concretely, we define
\begin{align}
v_t(x) = \frac{\partial s(h(z_f),t)}{\partial z_f}.
\end{align}
This gradient indicates that a small perturbation of $z_f$ along $v_t(x)$ increases $s(h(z_f),t)$
(e.g., for the feature $z_f$ of a \textit{brambling} image and the text $t$ = ``orange plumage,''
it changes $z_f$ such that its aligned embedding $h(z_f)$ moves toward the text embedding of ``orange plumage'' in the joint embedding space),
and thus $v_t(x)$ can be interpreted as a text-induced feature-change direction in the classifier feature space $Z_f$ (see \cref{sec:viz_direction_vector}).
In particular, 
when the aligner is affine, $h(z_f)=Wz_f+b$, 
the direction vector $v_t(x)=\partial s(h(z_f),t)/\partial z_f$ admits the following closed form:
\begin{align}
v_t(x) = W^{\top}\frac{1}{\|W z_f + b\|_2}
\left(\hat z_{\text{text}} - s(h(z_f),t)\,\frac{W z_f + b}{\|W z_f + b\|_2}\right),
\label{eq:vt_closed_form}
\end{align}
where $\hat z_{\text{text}} = \frac{E_{\text{text}}(t)}{\|E_{\text{text}}(t)\|_2}$.
The derivation is provided in Appendix~\ref{sec:deviation_vt}.
In practice, the magnitude of $v_t(x)$ only rescales the influence score in \cref{eq:faithfulness}, 
so we use the $\ell_2$-normalized direction $\hat{v}_t(x) = \frac{v_t(x)}{\|v_t(x)\|_2}$.

\par
Next, we compute the influence score.
For a sufficiently small $\epsilon$, 
by the first-order Taylor expansion of $g_c$ around $z_f$, 
we have
\begin{align}
g_c(z_f+\epsilon \hat{v}_t(x)) \approx g_c(z_f) + \epsilon \,\nabla_{z_f} g_c(z_f)^{\top} \hat{v}_t(x).
\label{eq:taylor_1st}
\end{align}
Substituting \cref{eq:taylor_1st} into \cref{eq:faithfulness} yields
\begin{align}
I_c(x; \hat{v}_t(x)) \approx \epsilon \,\nabla_{z_f} g_c(z_f)^{\top} \hat{v}_t(x).
\label{eq:influence_taylor}
\end{align}
The term $\nabla_{z_f} g_c(z_f)^{\top} \hat{v}_t(x)$ is exactly the directional derivative of $g_c$ at $z_f$
along the direction $\hat{v}_t(x)$, i.e.,
\begin{align}
\nabla_{z_f} g_c(z_f)^{\top} \hat{v}_t(x)
= \lim_{\delta\to 0}\frac{g_c(z_f+\delta \hat{v}_t(x)) - g_c(z_f)}{\delta}.
\label{eq:dirder_def}
\end{align}
Therefore, to obtain highly faithful explanations, we prefer texts whose induced direction 
$\hat{v}_t(x)$ yields a large positive directional derivative of the class-$c$ logit, i.e., 
$\nabla_{z_f} g_c(z_f)^{\top} \hat{v}_t(x)$.

\subsubsection{Textual explanation generation}
\label{sec:textual_explanation_generation}
We now describe how to produce textual explanations based on the influence score.
Let $\mathcal{B}(x,c)$ be a set of candidate textual descriptions for input $x$
and class $c$ (concept bank; see the next paragraph).
For each candidate $t \in \mathcal{B}(x,c)$, 
we compute the influence score $I_c(x; \hat v_t(x))$ via the directional derivative
$\nabla_{z_f} g_c(z_f)^{\top} \hat v_t(x)$.
We then rank all candidates by $I_c(x; \hat v_t(x))$ and output the top-$k$
texts with the largest positive scores as the textual explanation for the
prediction of class $c$ on input $x$.

\paragraph{Construction of concept bank.} 
The concept bank $\mathcal{B}(x,c)$ is a set of candidate concept descriptions
that are likely to be relevant to the prediction of class $c$ for input $x$.
Following~\cite{texter}, 
we construct $\mathcal{B}(x,c)$ using both a large language model (LLM) and a vision--language model (VLM):
the LLM is prompted with the class label $c$ to generate generic class-level descriptions, 
and the VLM is prompted with $(x,c)$ to obtain visually grounded descriptions.
In our implementation, 
we use \texttt{GPT\mbox{-}3.5\mbox{-}turbo}~\cite{gpt3.5} as the LLM and
\texttt{Qwen2.5\mbox{-}VL\mbox{-}7B\mbox{-}Instruct}~\cite{Qwen-VL,Qwen2VL,qwen2.5-VL} as the VLM;
further details are provided in Appendix~\ref{sec:detail_concept_bank}.

\begin{table*}[t]
\centering
\caption{Directional score results for the top-1/top-3/top-5 settings. ``Mean'' and ``NR'' denote the average directional score and its negative rate over all samples, respectively. Best values are in bold.}
\label{tab:direction_score}
\begin{tabular}{llcc|cc|cc}
\toprule
\multirow{2}{*}{\textbf{Model}} & \multirow{2}{*}{\textbf{Method}}
& \multicolumn{2}{c}{\textbf{Top1}}
& \multicolumn{2}{c}{\textbf{Top3}}
& \multicolumn{2}{c}{\textbf{Top5}} \\
\cmidrule(lr){3-4}\cmidrule(lr){5-6}\cmidrule(lr){7-8}
&& \multicolumn{1}{c}{\textbf{Mean} $\uparrow$} &\multicolumn{1}{c}{\textbf{NR} $\downarrow$} & \multicolumn{1}{c}{\textbf{Mean} $\uparrow$} & \multicolumn{1}{c}{\textbf{NR} $\downarrow$} & \multicolumn{1}{c}{\textbf{Mean}$ \uparrow$} & \multicolumn{1}{c}{\textbf{NR} $\downarrow$} \\
\midrule

\multirow{4}{*}{ResNet-18}
& Random  & -0.088 & 0.830  & -0.089 & 0.819 & 0.086 & 0.815 \\
& Text-To-Concept  &  0.024 & 0.438 &  0.0124 & 0.478 & 0.006 & 0.500 \\
& TEXTER  &  0.062 & 0.309 &  0.046 & 0.342 & 0.034 & 0.388 \\
& \cellcolor{gray!12}\mymethod & \cellcolor{gray!12}\textbf{0.161} &\cellcolor{gray!12} \textbf{0.037} & \cellcolor{gray!12}\textbf{0.132} &\cellcolor{gray!12} \textbf{0.064} & \cellcolor{gray!12}\textbf{0.114} & \cellcolor{gray!12}\textbf{0.087} \\
\midrule

\multirow{4}{*}{ResNet-50}
& Random  & -0.041  & 0.855 & -0.040 & 0.834 & -0.041 & 0.843 \\
& Text-To-Concept  &  0.014 & 0.390  &  0.007 & 0.464 &  0.003 & 0.495 \\
& TEXTER  &  0.031 & 0.254 &  0.021 & 0.335 &  0.016 & 0.374 \\
& \cellcolor{gray!12}\mymethod & \cellcolor{gray!12}\textbf{0.060} & \cellcolor{gray!12}\textbf{0.036} & \cellcolor{gray!12}\textbf{0.049} & \cellcolor{gray!12}\textbf{0.078} & \cellcolor{gray!12}\textbf{0.042} & \cellcolor{gray!12}\textbf{0.113} \\
\midrule

\multirow{4}{*}{DINO ResNet-50}
& Random  & -0.598 & 0.767 & -0.579 & 0.768 & -0.591 & 0.776 \\
& Text-To-Concept  &  0.328 & 0.354 &  0.237 & 0.406 &  0.185 & 0.428 \\
& TEXTER &  0.460 & 0.299 &  0.393 & 0.326 &  0.320 & 0.353 \\
& \cellcolor{gray!12}\mymethod & \cellcolor{gray!12}\textbf{1.425} & \cellcolor{gray!12}\textbf{0.023} & \cellcolor{gray!12}\textbf{1.183} & \cellcolor{gray!12}\textbf{0.045} & \cellcolor{gray!12}\textbf{1.041} & \cellcolor{gray!12}\textbf{0.069} \\
\midrule

\multirow{4}{*}{ViT}
& Random  & -0.014 & 0.621 &  0.015 & 0.631 & -0.015 & 0.630 \\
& Text-To-Concept  &  0.024 & 0.421 &  0.019 & 0.434 &  0.017 & 0.454 \\
& TEXTER &  0.056 & 0.274 &  0.045 & 0.313 &  0.040 & 0.338 \\
& \cellcolor{gray!12}\mymethod & \cellcolor{gray!12}\textbf{0.138} & \cellcolor{gray!12}\textbf{0.007} & \cellcolor{gray!12}\textbf{0.122} & \cellcolor{gray!12}\textbf{0.013} & \cellcolor{gray!12}\textbf{0.113} & \cellcolor{gray!12}\textbf{0.020} \\
\midrule

\multirow{4}{*}{Dino ViT-S/8}
& Random  & -0.027 & 0.787 & -0.025 & 0.783 & -0.026 & 0.806 \\
& Text-To-Concept  &  0.003 & 0.487 & -0.001 & 0.527 & -0.003 & 0.559 \\
& TEXTER  & -0.008 & 0.625 & -0.009 & 0.634 & -0.010 & 0.644 \\
& \cellcolor{gray!12}\mymethod & \cellcolor{gray!12}\textbf{0.047} & \cellcolor{gray!12}\textbf{0.042} & \cellcolor{gray!12}\textbf{0.038} & \cellcolor{gray!12}\textbf{0.069} & \cellcolor{gray!12}\textbf{0.033} & \cellcolor{gray!12}\textbf{0.096} \\
\bottomrule
\end{tabular}
\end{table*}

\subsection{Influence-based evaluation metric}
\label{sec:evaluation_metric}
We propose quantitative evaluation metrics for the faithfulness of textual explanations based on the influence score.
Specifically, we introduce two metrics.
The first metric,
which we call the \emph{directional score}, 
measures the local influence, i.e., 
how strongly an infinitesimal perturbation along a text-induced direction changes the class logit.
The second metric, referred to as the \emph{influence curve}, 
captures neighborhood-level influence under finite feature-space interventions along the same direction by aggregating influence scores over a sequence of steps in the feature space.
In the following, we describe these two metrics.

\paragraph{Directional score.}
For a given input $x$, class $c$, and a produced textual explanation $t'$,
we quantify its local influence by the directional derivative of the class-$c$
logit along the normalized text-induced direction $\hat v_{t'}(x)$ in
\cref{eq:dirder_def}, which we refer to as the \emph{directional score}.
A larger positive directional score indicates that the explanation $t'$ locally supports the prediction of class~$c$ and is therefore regarded as more faithful.

\paragraph{Influence curve.}
While the directional score focuses on infinitesimal perturbations, 
we also evaluate how a produced explanation $t'$ affects the prediction when the representation is moved by a finite amount along its direction.
Intuitively, 
a faithful explanation should make the class-$c$ logit increase when we move the feature toward $\hat v_{t'}(x)$ and decrease when we move it in the opposite direction. 
Moreover, the magnitude of these changes should be larger for more faithful explanations.

\par
Let $\{\alpha_k\}_{k=1}^K$ with $\alpha_k > 0$ be a set of step sizes.
For each $\alpha_k$, 
we compute the logit changes
\begin{align}
\Delta_c^{\mathrm{ins}}(\alpha_k) &= g_c\bigl(z_f + \alpha_k \hat v_{t'}(x)\bigr) - g_c(z_f) && \text{(insertion)}, \nonumber \\
\Delta_c^{\mathrm{del}}(\alpha_k) &= g_c(z_f) - g_c\bigl(z_f - \alpha_k \hat v_{t'}(x)\bigr) && \text{(deletion)}.
\label{eq:influence_curve_deltas}
\end{align}
Here, $\Delta_c^{\mathrm{ins}}(\alpha_k)$ measures how much the class-$c$ evidence increases when the representation is moved toward the explanation direction (insertion), 
and $\Delta_c^{\mathrm{del}}(\alpha_k)$ measures how much it decreases when the representation is moved in the opposite direction (deletion).
To avoid perturbations that move the features too far from the typical data
distribution, 
we control the intervention strength by choosing the step size relative to the feature norm and set $\alpha_k = \rho_k \,\|z_f\|_2$ with small relative steps $\rho_k$.
For both insertion and deletion, 
larger positive values indicate stronger changes in the class-$c$ score at each step, which we interpret as higher faithfulness of the explanation $t'$.

\begin{figure*}[t]
  \centering
  \includegraphics[width=\linewidth]{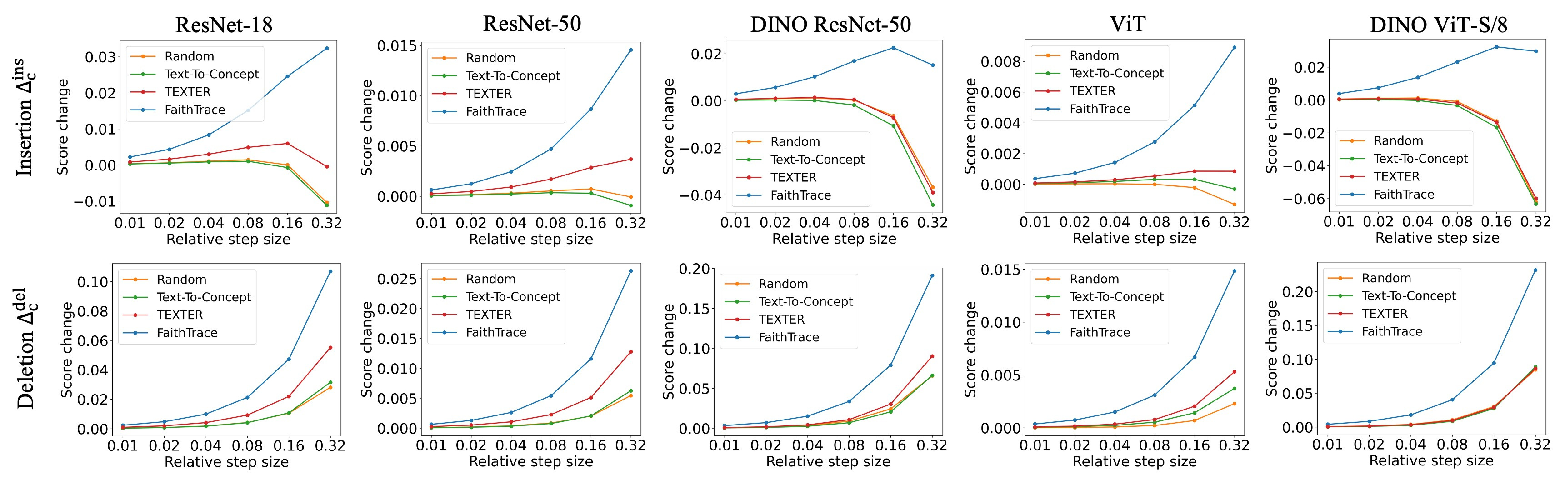}
  \caption{Influence curves for the insertion (first row) and the deletion (second row) under the top-3 setting. At each step, larger positive values indicate higher faithfulness.}
  \label{fig:ins_del}
\end{figure*}

\par
In practice, instead of using the raw logit, 
we evaluate these changes on a margin-based confidence score.
We define the margin
\begin{align}
m_c(z_f) &= g_c(z_f) - \max_{j \neq c} g_j(z_f),
\end{align}
and map it to
\begin{align}
q_c(z_f) &= \sigma\bigl(m_c(z_f)\bigr) \in (0,1),
\end{align}
where $\sigma(\cdot)$ is the sigmoid function.
This converts the advantage of class $c$ over the others into a normalized confidence, making influence scores more comparable across examples.
In our experiments, we compute the insertion and deletion changes by replacing $g_c(\cdot)$ with $q_c(\cdot)$ in \cref{eq:influence_curve_deltas}.

\section{Experiments}
\label{sec:experiments}

We evaluate \mymethod on both CNN- and Transformer-based image classifiers.
For CNNs, 
we use ResNet-18, ResNet-50~\cite{resnet}, and a DINO-pretrained ResNet-50~\cite{dino};
for Transformers, 
we use ViT~\cite{vit} and DINO ViT-S/8~\cite{dino}.
The aligner $h$ projects classifier features into the CLIP image-embedding space defined by the ViT-B/16 vision encoder.
Following Text-To-Concept~\cite{text_to_concept}, we train $h$ on a subset of ImageNet-1K~\cite{imagenet}, using 20\,\% of the training split as $\mathcal{D}_{\text{train}}$ and adopting the same data split and training protocol.
When pretrained weights for the aligner are publicly available, 
we use the official checkpoints released by the authors\footnote{\url{https://github.com/k1rezaei/Text-to-concept}};
otherwise, we train the aligner with the default settings of~\cite{text_to_concept}.
In our experiments, 
the concept bank $\mathcal{B}(x,c)$ contains 130 descriptions per prediction (100 LLM-generated and 30 VLM-generated), 
and all evaluations are conducted on the ImageNet-1K test set.

\par
As baselines, we consider three methods.
The first one, denoted as \textbf{Random}, 
selects $K$ candidate descriptions uniformly at random from the concept bank $\mathcal{B}(x,c)$ and outputs them
as the explanation.
Since $\mathcal{B}(x,c)$ consists of LLM/VLM-generated candidate concepts intended to be relevant to the prediction,
Random serves as a baseline that uses these prediction-relevant descriptions without any further selection or ranking.
The second baseline, \textbf{Text-To-Concept}~\cite{text_to_concept}, 
follows~\cref{eq:sim_score_def_short} and measures the cosine similarity between the (global) aligned image feature $h(z_f)$ and each candidate text in the joint vision-language space, 
retrieving the top-$k$ descriptions with the highest similarity scores.
The third baseline, \textbf{TEXTER}~\cite{texter}, instead compares each candidate text with an isolated concept image that visualizes decision-critical features: 
in \cref{eq:sim_score_def_short}, the original image $x$ is replaced by the corresponding feature-visualized concept image, 
and the top-$k$ descriptions are retrieved by similarity.
For TEXTER, we follow the original configuration and use Integrated Gradients to select six important neurons and enable the sparse autoencoder option when generating concept images~\cite{text_to_concept}.

\begin{table*}[t]
\centering
\caption{Sums over all steps in the influence curves for the insertion (ins) and the deletion (del) under the top-3 setting.
Higher is better. Best values are in bold.
All values are reported in units of $10^{-1}$ for readability.
}
\label{tab:influence_curve}
\begin{tabular}{llcccc}
\toprule
\textbf{Model} & \textbf{Metric }
& {\small\textbf{Random }}
& {\small\textbf{Text-To-Concept }}
& {\small\textbf{TEXTER }}
& {\cellcolor{gray!12}\small\textbf{\mymethod}} \\
\midrule
\multirow{2}{*}{ResNet-18}
& Ins  $\uparrow$& -0.069 & -0.092 & 0.159 & \cellcolor{gray!12}\textbf{0.870} \\
& Del $\uparrow$& 0.458  & 0.492  & 0.933 & \cellcolor{gray!12}\textbf{1.925} \\
\midrule
\multirow{2}{*}{ResNet-50}
& Ins $\uparrow$& 0.018 & 0.002 & 0.099 & \cellcolor{gray!12}\textbf{0.323} \\
& Del $\uparrow$& 0.090 & 0.096 & 0.220 & \cellcolor{gray!12}\textbf{0.479} \\
\midrule
\multirow{2}{*}{DINO ResNet-50}
& Ins $\uparrow$& -0.401 & -0.559 & -0.424 &\cellcolor{gray!12} \textbf{0.734} \\
& Del $\uparrow$& 1.043  & 0.975  & 1.379 &\cellcolor{gray!12} \textbf{3.290} \\
\midrule
\multirow{2}{*}{ViT}
& Ins $\uparrow$& -0.015 & 0.007 & 0.028 & \cellcolor{gray!12}\textbf{0.193} \\
& Del $\uparrow$& 0.033  & 0.061 & 0.087 & \cellcolor{gray!12}\textbf{0.272} \\
\midrule
\multirow{2}{*}{DINO ViT-S/8}
& Ins $\uparrow$& -0.732 & -0.832 & -0.737 & \cellcolor{gray!12}\textbf{1.103} \\
& Del $\uparrow$& 1.317  & 1.286  & 1.304 & \cellcolor{gray!12}\textbf{3.974} \\
\bottomrule
\end{tabular}
\end{table*}

\begin{figure*}[t]
  \centering
  \includegraphics[width=0.9\linewidth]{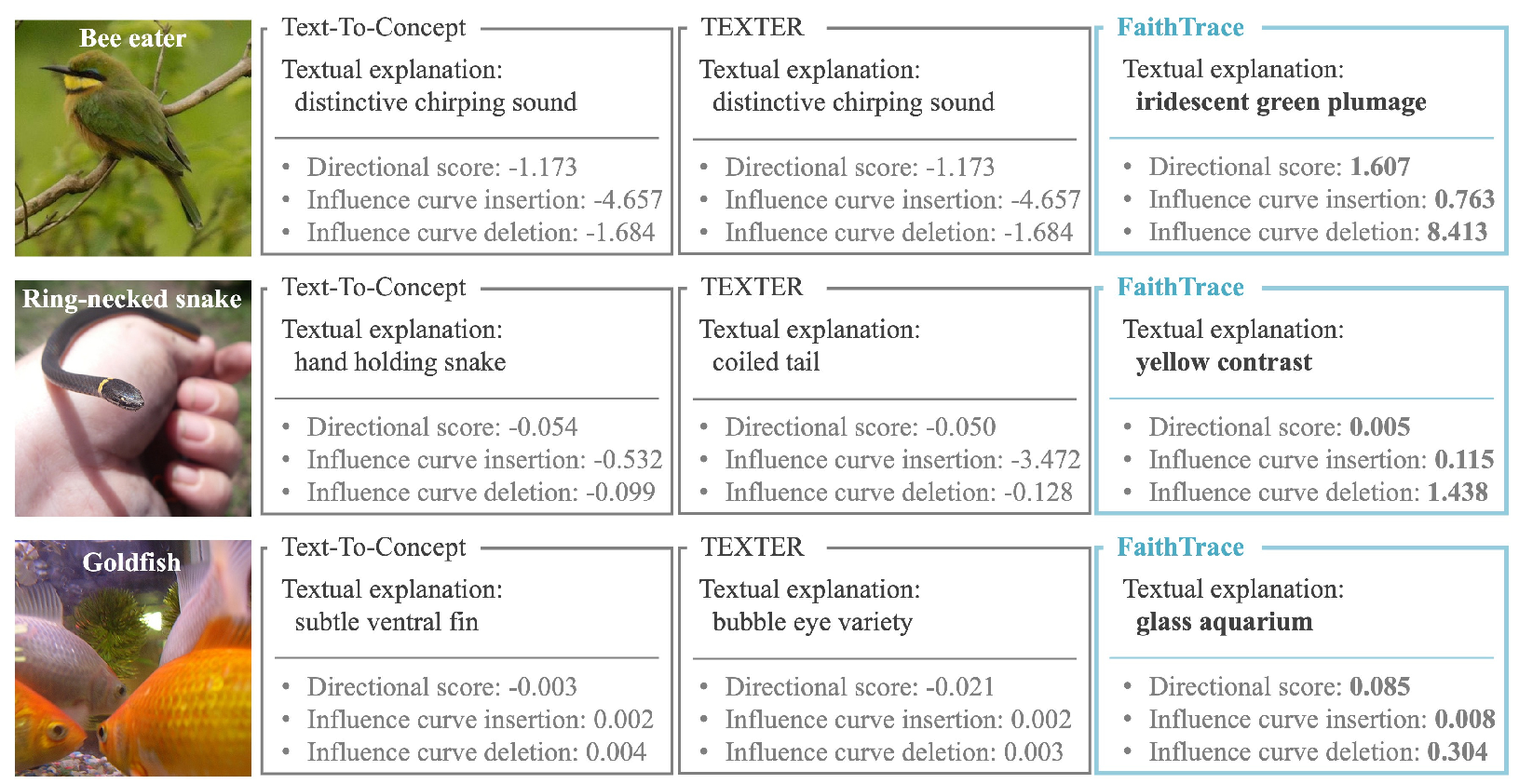}
  \caption{Comparison of top-1 textual explanations produced by each method.
  For each original image, we show the predicted label.
  In each example, we also report the directional score and the summed insertion
and deletion influence curve scores (in units of $10^{-1}$ for readability).}
  \label{fig:compare}
\end{figure*}

\begin{table*}[t]
\centering
\caption{Quantitative evaluation results for misclassification cases.
We report the directional score (Mean/NR) and the influence curve sums (Ins/Del, in units of $10^{-1}$) under the top-3 setting.
Best values are shown in bold.
}
\label{tab:miss_quantitative}
\begin{tabular}{llcccc}
\toprule
\textbf{Model} & \textbf{Method} &
\multicolumn{2}{c}{\small\textbf{Directional score}} &
\multicolumn{2}{c}{\footnotesize\textbf{Influence curve}} \\
\cmidrule(lr){3-4}\cmidrule(lr){5-6}
& &
\textbf{Mean} $\uparrow$ & \textbf{NR} $\downarrow$ &
\textbf{Ins} $\uparrow$ & \textbf{Del} $\uparrow$ \\
\midrule
\multirow{4}{*}{ResNet-18}
& Random          & -0.043 & 0.647 &  0.093 & 1.575 \\
& Text-To-Concept &  0.053 & 0.290 & -0.172 & 1.516 \\
& TEXTER          &  0.068 & 0.277 &  0.579 & 2.571 \\
& \cellcolor{gray!12}\mymethod       & \cellcolor{gray!12}\textbf{0.190} & \cellcolor{gray!12}\textbf{0.000} & \cellcolor{gray!12}\textbf{2.496} & \cellcolor{gray!12}\textbf{4.869} \\
\midrule
\multirow{4}{*}{ResNet-50}
& Random          & -0.027 & 0.713 & 0.240 & 0.434 \\
& Text-To-Concept &  0.021 & 0.253 & 0.253 & 0.477 \\
& TEXTER          &  0.035 & 0.183 & 0.672 & 0.947 \\
& \cellcolor{gray!12}\mymethod       & \cellcolor{gray!12}\textbf{0.069} & \cellcolor{gray!12}\textbf{0.007} & \cellcolor{gray!12}\textbf{1.375} & \cellcolor{gray!12}\textbf{1.732} \\
\midrule
\multirow{4}{*}{DINO ResNet-50}
& Random          & -0.243 & 0.593 & -0.001 & 3.283 \\
& Text-To-Concept &  0.502 & 0.263 & -0.024 & 3.118 \\
& TEXTER          &  0.668 & 0.220 & -0.002 & 3.440 \\
& \cellcolor{gray!12}\mymethod       & \cellcolor{gray!12}\textbf{1.700} & \cellcolor{gray!12}\textbf{0.000} & \cellcolor{gray!12}\textbf{2.971} & \cellcolor{gray!12}\textbf{8.140} \\
\midrule
\multirow{4}{*}{ViT}
& Random          & -0.028 & 0.657 & 0.187 & 0.293 \\
& Text-To-Concept &  0.043 & 0.267 & 0.184 & 0.303 \\
& TEXTER          &  0.043 & 0.283 & 0.389 & 0.537 \\
& \cellcolor{gray!12}\mymethod       & \cellcolor{gray!12}\textbf{0.130} & \cellcolor{gray!12}\textbf{0.000} & \cellcolor{gray!12}\textbf{0.874} & \cellcolor{gray!12}\textbf{1.077} \\
\midrule
\multirow{4}{*}{DINO ViT-S/8}
& Random          & -0.019 & 0.737 & -1.752 & 2.633 \\
& Text-To-Concept &  0.002 & 0.497 & -2.069 & 2.989 \\
& TEXTER          & -0.010 & 0.623 & -1.961 & 2.405 \\
& \cellcolor{gray!12}\mymethod       & \cellcolor{gray!12}\textbf{0.044} & \cellcolor{gray!12}\textbf{0.020} & \cellcolor{gray!12}\textbf{3.108} & \cellcolor{gray!12}\textbf{7.751} \\
\bottomrule
\end{tabular}
\end{table*}

\begin{figure}
  \centering
  \includegraphics[width=\linewidth]{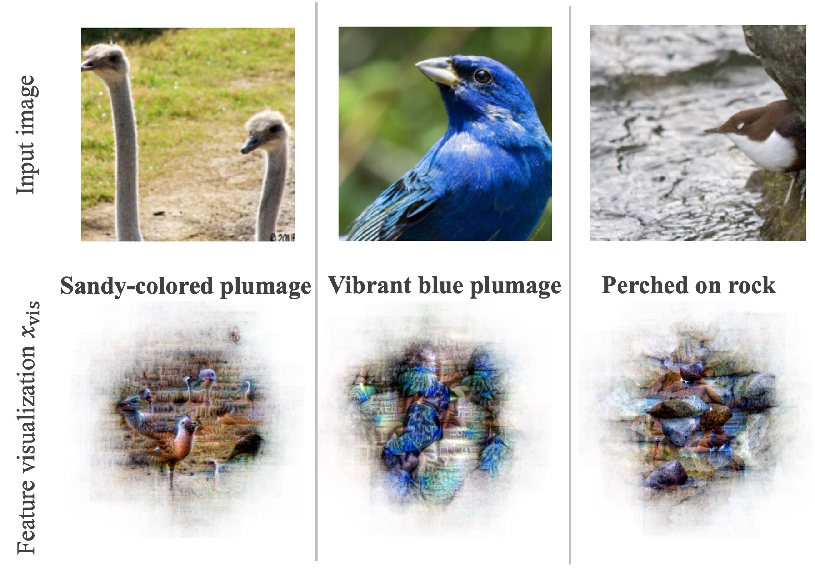}
  \caption{Examples of input images, produced textual explanations, and the corresponding
  feature visualizations obtained by applying MACO~\cite{maco}.}
  \label{fig:fv}
\end{figure}

\begin{figure*}[t]
  \centering
  \includegraphics[width=\linewidth]{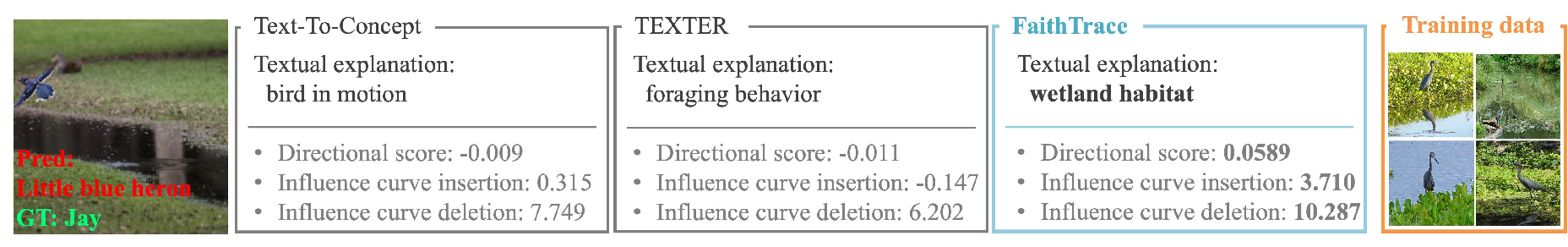}
  \caption{
    Comparison of top-1 textual explanations from each method on an image where a \textit{jay} is misclassified as a \textit{little blue heron}; each explanation is annotated with its directional score and summed insertion/deletion influence curve (in units of $10^{-1}$). 
    The rightmost column shows example training images of the \textit{little blue heron} class.
    }
  \label{fig:compare_miss}
\end{figure*}

\subsection{Quantitative evaluations for explanations}
\label{sec:quantitative_evaluations}

We quantitatively evaluate the faithfulness of explanations from our method and baselines using the directional score and influence curve (see \cref{sec:evaluation_metric}).
In the following, we describe each result in detail.

\paragraph{Directional score results.}
Table~\ref{tab:direction_score} reports the directional score results. 
For each method, 
we present the mean directional score (Mean) and the negative rate (NR), i.e., the fraction of cases where the directional score is negative, 
for explanations retrieved in the top-1/top-3/top-5 settings. 
We evaluate on 1,000 images sampled from the ImageNet-1K test set (200 classes, 5 images per class). 
Higher Mean and lower NR indicate better faithfulness.

\par
Table~\ref{tab:direction_score} shows that, 
across all models and under all retrieval settings (top-1/top-3/top-5), 
the proposed method consistently outperforms the baselines by a large margin. 
In the baselines, the Mean can become negative, 
indicating that the retrieved explanations correspond to directions that do not increase the class-$c$ logit but instead decrease it (i.e., they do not support the prediction). 
In contrast, 
the proposed method achieves clearly higher Mean values and substantially lower NR. 
These results suggest that, compared to the baselines, 
our method stably produces more faithful explanations for the classifier’s prediction.
Note that NR does not become exactly zero even for our method because the concept bank $\mathcal{B}(x,c)$ is finite and may not always contain concepts that match the decision-relevant evidence for a given input $x$. Improving the coverage and relevance of concept banks remains a common challenge in this research area.

\paragraph{Influence curve results.}
As described in \cref{sec:evaluation_metric},
we compute the insertion and deletion curves along the explanation direction,
replacing the raw logit $g_c(\cdot)$ with the margin-based confidence score $q_c(\cdot)$.
To control the intervention strength, we sweep the relative step size $\rho \in \{0.01, 0.02, 0.04, 0.08, 0.16, 0.32\}$ and set $\alpha_k = \rho_k \|z_f\|_2$ at each step.
For all methods, we evaluate the influence curve under the top-3 retrieval setting on 1{,}000 images sampled from the ImageNet-1K test set (200 classes, 5 images per class).

\par
Figure~\ref{fig:ins_del} shows the insertion and deletion curves,
where each point is averaged over all images and the top-3 retrieved explanations.
At each step, larger positive insertion and deletion values indicate higher faithfulness, 
as they reflect stronger changes in favor of class $c$ along the explanation direction and its opposite.
Across all step sizes, our method consistently exhibits larger changes than the baselines, indicating that it produces more faithful explanations.
In contrast, some baselines yield negative insertion values, 
meaning that moving along the retrieved explanation direction can even reduce the margin-based confidence $q_c$ of class $c$.
This behavior suggests that the retrieved texts are not sufficiently aligned with the decision-relevant evidence of the classifier.
As the summary,
\cref{tab:influence_curve} reports the sums over all steps in the insertion and deletion curves,
where higher values are better.
The proposed method consistently outperforms the baselines by large margins,
indicating its effectiveness.
More results are in Appendix.~\ref{sec:more_ins_del}.

\subsection{Qualitative results}
\label{sec:qaualitative_evaluations}
Figure~\ref{fig:compare} compares the top-1 textual explanations produced by Text-To-Concept, TEXTER, and the proposed method.
For each example, 
we also display the directional score and the summed influence curve values for the insertion and the deletion.
As shown in \cref{fig:compare}, 
the proposed method tends to produce more faithful explanations that better drive the prediction than the baselines.
In the first row (``bee eater''), 
the baselines refer to a phrase such as ``distinctive chirping sound,''
which is not clearly grounded in the image,
whereas our method focuses on ``iridescent green plumage,''
which matches the high faithfulness scores reported by our metrics.
In the second row (``ring-necked snake''), 
Text-To-Concept is distracted by the human hand in the image, 
and TEXTER outputs a generic description of snakes that mentions attributes not actually visible, 
while our method correctly highlights the ``yellow contrast'' around the neck as a discriminative
feature.
In the third row (``goldfish''), the baseline explanations sound plausible but receive low faithfulness scores, whereas our method identifies the ``glass aquarium'' in the background as the actual driver of the prediction,
which is consistent with its high scores.
More results are in Appendix.~\ref{sec:more_compare}.

\subsection{Visualizing text-induced direction vectors}
\label{sec:viz_direction_vector}
Our method quantifies logit changes when the classifier feature $z_f$ is moved along the normalized direction $\hat v_t(x)$ in \cref{eq:vt_closed_form}.
To verify that this movement enhances the visual cues described by the text, 
we synthesize images via feature visualization~\cite{fv,nguyen16,nguyen17,maco,vital}.
Concretely, we use magnitude-constrained optimization (MACO)~\cite{maco} to match the feature representation to $z_{\text{target}} = z_f + \alpha \hat v_t(x)$.
We obtain the visualization image $x_{\text{vis}}$ by solving
\begin{align}
x_{\text{vis}} 
= \underset{x' \in \mathcal{X}}{\operatorname{argmin}}\,
\bigl\|f(x') - z_{\text{target}}\bigr\|_2^2,
\label{eq:maco}
\end{align}
where $\mathcal{X}$ denotes the input image space.
We set $\alpha = 100$ so that the shift along the
text-induced direction is visually noticeable while keeping the optimization stable.
\par
Figure~\ref{fig:fv} shows the input images, the produced textual explanations,
and the corresponding feature visualizations obtained by applying MACO following \cref{eq:maco}.
For example, in the first and second columns, 
the explanations ``sandy-colored plumage'' and ``vibrant blue plumage'' yield visualizations in
which the related colors and feather-like structures are clearly emphasized.
In the third column (``perched on rock''), 
the synthesized image highlights rock-like textures.
These results indicate that moving along the text-induced direction enhances visual cues consistent with the corresponding text.

\subsection{Evaluation on misclassification cases}
\label{sec:eval_miss}
Faithful explanations on misclassified samples are particularly valuable for diagnosing biases hidden in the data.
In this experiment, 
we focus on misclassification cases and conduct both quantitative and qualitative evaluations.

\par
Table~\ref{tab:miss_quantitative} reports quantitative metrics for the directional score (mean and the proportion of negative scores) and the influence curve (insertion/deletion sums), 
computed on 100 misclassified images per model under the top-3 setting.
Across all models, 
the proposed method consistently outperforms the baselines, 
indicating that the provided explanations better follow the model's decision behavior. 
More results are provided in Appendix~\ref{sec:more_miss}.

\par
Figure~\ref{fig:compare_miss} shows a misclassification example where a \textit{jay} is predicted as a \textit{little blue heron}.
Baselines output generic phrases such as ``bird in motion'' and ``foraging behavior,''
which could describe many bird images and do not clarify this error.
In contrast, our method highlights context such as ``wetland habitat,'' 
which is consistent with the fact that most \textit{little blue heron} training images depict birds in wetlands.
This suggests that the prediction is driven by a spurious background correlation and that our explanations expose such dataset biases.
Overall, our method provides more faithful and diagnostically useful explanations even for misclassification cases.

\section*{Acknowledgments}
This work was supported by JSPS KAKENHI Grant Numbers JP23K24914,
and 26K02967,
JST PRESTO Grant Number JPMJPR24K4,
JST BOOST Program Grant Number JPMJBY24C6,
and ROIS NII Open Collaborative Research 261S07-24168.


\clearpage
\setcounter{page}{1}
\maketitlesupplementary

\appendix

\renewcommand{\thesection}{\Alph{section}}

\section{Additional discussion of related work}
\label{sec::more_releted_works}

In this section, 
we provide additional details on concept bottleneck models, 
natural language explanation models, 
and LVLM-based reasoning systems that could not be covered in the main text due to space constraints.

\subsection{Concept bottleneck models}
\label{sec:related_works_cbm}
Concept-based methods such as Network Dissection~\cite{8099837} and TCAV~\cite{tcav} analyze how internal representations correlate with human-understandable concepts.
More recently, Concept Bottleneck Models (CBMs)~\cite{cbn, label_free_cbn, labo, incremental_cbn, language_cbn} insert an intermediate concept layer that predicts human-defined concepts before the final label, improving interpretability but requiring concept supervision and model retraining, even though extensions mitigate this cost using CLIP-based pseudo-labeling~\cite{label_free_cbn} or LLM-generated concepts~\cite{hybrid_cbn}.
In contrast, we consider standard classifiers trained only with images and class labels, and generate textual explanations without modifying or retraining the backbone, bringing concept-level interpretability to existing pretrained models in a zero-shot manner.

\subsection{Natural language explanation (NLE) models}
\label{sec:related_works_nle}
NLE models learn to verbalize the reasons behind a vision or vision--language model's prediction~\cite{generate_visual_explanation, park_nle, nlx_gpt, sharma, debil}. 
Typical approaches couple a task network with a language model (e.g., GPT-2~\cite{gpt2}), 
as in NLX-GPT~\cite{nlx_gpt} and Uni-NLX~\cite{uni_nlx}, to generate explanations. 
However, they require paired prediction–explanation annotations and tend to mirror annotation bias rather than the model's actual decision process~\cite{zsnle}, 
whereas our method produces textual explanations for image classifiers in a zero-shot manner.
Large vision–language models (LVLMs)~\cite{pmlr-v162-li22n,NEURIPS2023_6dcf277e,openai2024gpt4technicalreport} can generate rich image descriptions and perform general visual reasoning, 
but they are optimized for generic vision–language understanding rather than for explaining the decision pathway of a specific classifier.

\section{Derivation of \cref{eq:vt_closed_form}}
\label{sec:deviation_vt}

In this section,
we derive the closed form of the text-induced direction vector
$v_t(x)=\partial s(h(z_f),t)/\partial z_f$ used in \cref{eq:vt_closed_form}.
Throughout, we treat gradients $\partial s/\partial z$ as column vectors.

Given an input image $x$,
the classifier feature is $z_f=f(x)\in\mathbb{R}^{d}$.
We use an affine aligner $h(z_f)=Wz_f+b\in\mathbb{R}^{m}$,
where $W\in\mathbb{R}^{m\times d}$ and $b\in\mathbb{R}^{m}$.
For a text description $t$,
the CLIP text embedding is $z_{\text{text}}=E_{\text{text}}(t)\in\mathbb{R}^{m}$,
and we define
\begin{align}
\hat z_h = \frac{h(z_f)}{\|h(z_f)\|_2},
\qquad
\hat z_{\text{text}} = \frac{z_{\text{text}}}{\|z_{\text{text}}\|_2}.
\end{align}
The CLIP similarity score is defined as
\begin{align}
s(h(z_f),t)=\hat z_h^{\top}\hat z_{\text{text}}.
\label{eq:app_sim_def}
\end{align}
For brevity, we write $s:=s(h(z_f),t)$ in what follows.

\par
Let $z_h=h(z_f)$.
Since $s$ is a scalar function of $z_h$,
by the chain rule,
\begin{align}
\frac{\partial s}{\partial z_f}
=
\left(\frac{\partial z_h}{\partial z_f}\right)^{\top}
\frac{\partial s}{\partial z_h}.
\label{eq:app_chain_rule}
\end{align}
In the following,
we first derive $\partial s/\partial z_h$,
then derive $\partial z_h/\partial z_f$,
and substitute both into \cref{eq:app_chain_rule}.

\par
Let $r(z_h)=\|z_h\|_2$ and $a(z_h)=\hat z_{\text{text}}^{\top} z_h$.
Then $\hat z_h = z_h/r(z_h)$ and
\begin{align}
s = \hat z_h^{\top}\hat z_{\text{text}}
=
\left(\frac{z_h}{r(z_h)}\right)^{\top}\hat z_{\text{text}}
=
\frac{a(z_h)}{r(z_h)}.
\label{eq:app_s_ar}
\end{align}
Differentiating \cref{eq:app_s_ar} w.r.t.\ $z_h$ (quotient rule) gives
\begin{align}
\frac{\partial s}{\partial z_h}
 = \frac{1}{r(z_h)}\frac{\partial a(z_h)}{\partial z_h}
-\frac{a(z_h)}{r(z_h)^2}\frac{\partial r(z_h)}{\partial z_h}.
\label{eq:app_quotient}
\end{align}
Here, $\frac{\partial a(z_h)}{\partial z_h}=\hat z_{\text{text}}$,
and since $r(z_h)=\|z_h\|_2=\sqrt{z_h^{\top}z_h}$,
\begin{align}
\frac{\partial r(z_h)}{\partial z_h}
=
\frac{z_h}{\|z_h\|_2}
=
\hat z_h,
\label{eq:app_dr}
\end{align}
we get
\begin{align}
\frac{\partial s}{\partial z_h}
 & = \frac{1}{r(z_h)}\hat z_{\text{text}}  - \frac{a(z_h)}{r(z_h)^2} \hat z_h\\
 & = \frac{1}{\|z_h\|_2}\hat z_{\text{text}}-\frac{a(z_h)}{\|z_h\|_2^2}\hat z_h.
\label{eq:app_ds_dzh_mid}
\end{align}
Using $a(z_h)=s\,\|z_h\|_2$ from \cref{eq:app_s_ar}, we obtain the compact form
\begin{align}
\frac{\partial s}{\partial z_h}
=
\frac{1}{\|z_h\|_2}\left(\hat z_{\text{text}} - s\,\hat z_h\right).
\label{eq:app_ds_dzh}
\end{align}
Next, we derive $\partial z_h/\partial z_f$.
For $z_h = W z_f + b$, the Jacobian is constant:
\begin{align}
\frac{\partial z_h}{\partial z_f}=W .
\label{eq:app_jacobian}
\end{align}
Finally, substituting \cref{eq:app_ds_dzh,eq:app_jacobian} into
\cref{eq:app_chain_rule} yields
\begin{align}
v_t(x)
=
\frac{\partial s}{\partial z_f}
=
W^{\top}\frac{1}{\|W z_f + b\|_2}
\left(
\hat z_{\text{text}} - s\,\frac{W z_f + b}{\|W z_f + b\|_2}
\right),
\end{align}
which matches \cref{eq:vt_closed_form}.

\section{Details of concept bank construction}
\label{sec:detail_concept_bank}
Following~\cite{texter}, 
we construct the concept bank $\mathcal{B}(x,c)$ by leveraging both an LLM and a VLM. 
We detail the prompts used for each model below.

\paragraph{LLM-based concept generation.}
We query the LLM to obtain concepts that characterize class $c$ at a category level, rather than being specific to an individual image.
These concepts represent attributes typically associated with the class.
To encourage concise, atomic concepts, we restrict each description to a short 1--3 word phrase, and we filter out trivial outputs that merely repeat the class name.
The LLM prompt includes the target class name (\{class\_name\}), 
the set of concepts collected so far  (\{existing\_concepts\}), 
and an illustrative question--answer example.
In one inference, the LLM is asked to produce 10 additional candidate descriptions.
After generation, we remove duplicates as a post-processing step.
The exact prompt is as follows:

\begin{promptverb}
Template variables (filled by the implementation as input variables):
- {class_name}: the target object class name.
- {existing_concepts}: already generated concepts.

Important guidelines for generating visual concepts:
1. Generate GENERAL concepts that can apply to many different photos of the same object type.
2. Include both OBJECT features (e.g., shape, color, parts) AND CONTEXT features (e.g., background, environment, setting).
3. Keep concepts short and specific (1-3 words).
4. DO NOT include class names or object names directly.

Q: What are useful visual features for distinguishing a lemur in a photo?
A: There are several useful visual features to tell there is a lemur in a photo:
- long tail
- large eyes
- gray fur
- trees
- branches
- forest

Q: What are useful features for distinguishing a {class_name} in a photo?
Already generated concepts (DO NOT repeat these): {existing_concepts}.
A: There are several useful visual features to tell there is a {class_name} in a photo. Generate approximately 10 visual concepts to provide comprehensive coverage:
\end{promptverb}

\paragraph{VLM-based concept generation.}
We use the VLM to obtain concepts that are explicitly grounded in the visual evidence of the input image $x$ for class $c$.
These image-specific concepts describe attributes that are observable in the particular instance and thus complement the class-level, image-agnostic concepts produced by the LLM.
As with the LLM, we restrict each concept to a compact 1--3 word phrase and discard outputs that merely repeat the class name.
In contrast to the LLM setting, the VLM additionally receives the image $x$ as visual input together with the text prompt below.
We otherwise follow the same generation protocol as in the LLM case, including the post-processing step (e.g., duplicate removal).

\begin{promptverb}
Template variables (filled by the implementation as input variables):
- {class_name}: the target object class name.
- {existing_concepts}: already generated concepts.

Important guidelines for generating visual concepts:
1. Generate DETAILED and SPECIFIC concepts that can apply to this image.
2. Include both OBJECT features (e.g., shape, color, parts) AND CONTEXT features (e.g., background, environment, setting).
3. Keep concepts short and specific (1-3 words).
4. DO NOT include class names or object names directly.

Examples:
Q: Look at this image carefully. Based on what you can actually see in the image, identify useful visual features that help distinguish this as a koi fish.
A: There are several useful visual features to tell there is a koi fish in a photo:
- bright orange scales
- curved tail fin
- spotted pattern
- long body
- pointed snout
- water surface

Q: Look at this image carefully. Based on what you can actually see in the image, identify useful visual features that help distinguish this as a {class_name}.
Already generated concepts (DO NOT repeat these): {existing_concepts}.
A: There are several useful visual features to tell there is a {class_name} in a photo. Generate approximately 10 visual concepts to provide comprehensive coverage:
\end{promptverb}

\begin{figure*}[t]
  \centering
  \includegraphics[width=\linewidth]{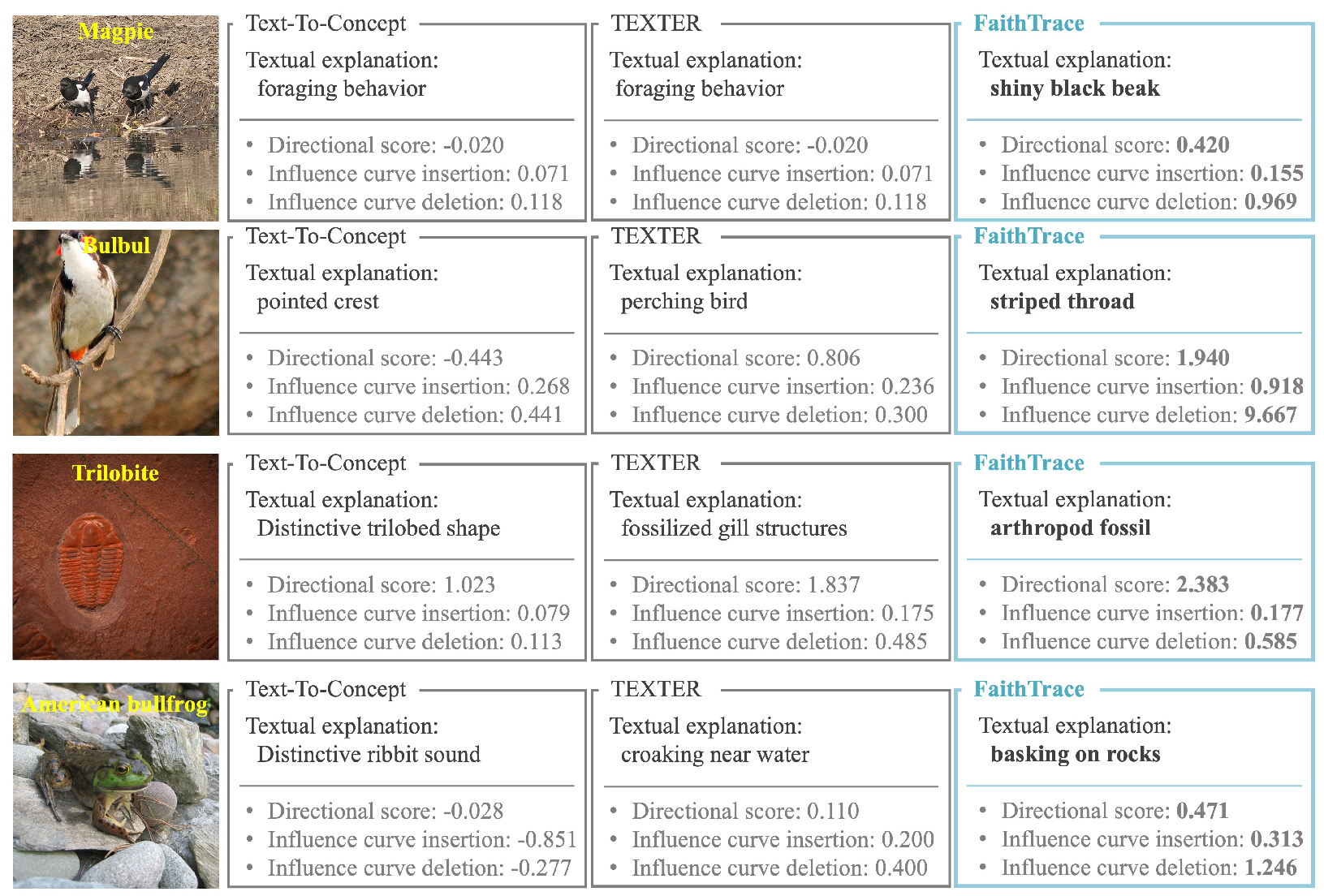}
  \caption{Comparison of top-1 textual explanations produced by each method.
  For each original image, we show the predicted label.
  In each example, we also report the directional score and the summed insertion
and deletion influence curve scores (in units of $10^{-1}$ for readability).}
  \label{fig:compare_more}
\end{figure*}

\section{More qualitative results}
\label{sec:more_compare}
Figure~\ref{fig:compare_more} presents a qualitative comparison by showing,
for each method,
the retrieved textual explanation together with its quantitative faithfulness measures:
the directional score and the influence curve sums for insertion and deletion.
Overall, the proposed method tends to select distinctive visual cues that are directly observable in the input,
and consequently achieves consistently higher directional score and influence curve values.

\par
For the magpie example (top row),
both Text-To-Concept and TEXTER output a generic,
behavior-oriented description (e.g., "foraging behavior"), 
which does not pinpoint a distinctive visual attribute. 
Accordingly, their directional score is negative (-0.020), 
indicating that the retrieved explanation direction does not increase the class-$c$ logit. 
In contrast, FaithTrace highlights a more discriminative appearance cue ("shiny black beak"),
yielding a substantially higher directional score (0.420) as well as larger insertion/deletion values (0.155 / 0.969).
This suggests that the selected explanation is better aligned with features that actually promote the prediction.

\par
For the trilobite example (third row),
TEXTER ("fossilized gill structures") and FaithTrace ("arthropod fossil") both capture the essential attribute that the instance is a fossil.
Consistent with this, their quantitative scores are broadly similar,
with nearly identical insertion values (0.175 vs. 0.177).
This case illustrates that when multiple methods recover a comparable,
visually grounded rationale,
the proposed metrics reflect that agreement.

\par
For the American bullfrog example (bottom row),
Text-To-Concept produces a description that is not directly verifiable from the image ("distinctive ribbit sound"), 
making the explanation less grounded in visual evidence. 
TEXTER also appears to be biased toward a stereotypical context ("croaking near water"),
even though the input image does not clearly depict water.
In contrast, FaithTrace outputs "basking on rocks,"
which matches the visible rocky background, 
and achieves a higher directional score (0.471) and larger insertion/deletion values (0.313 / 1.246).

\begin{figure*}[t]
  \centering
  \includegraphics[width=\linewidth]{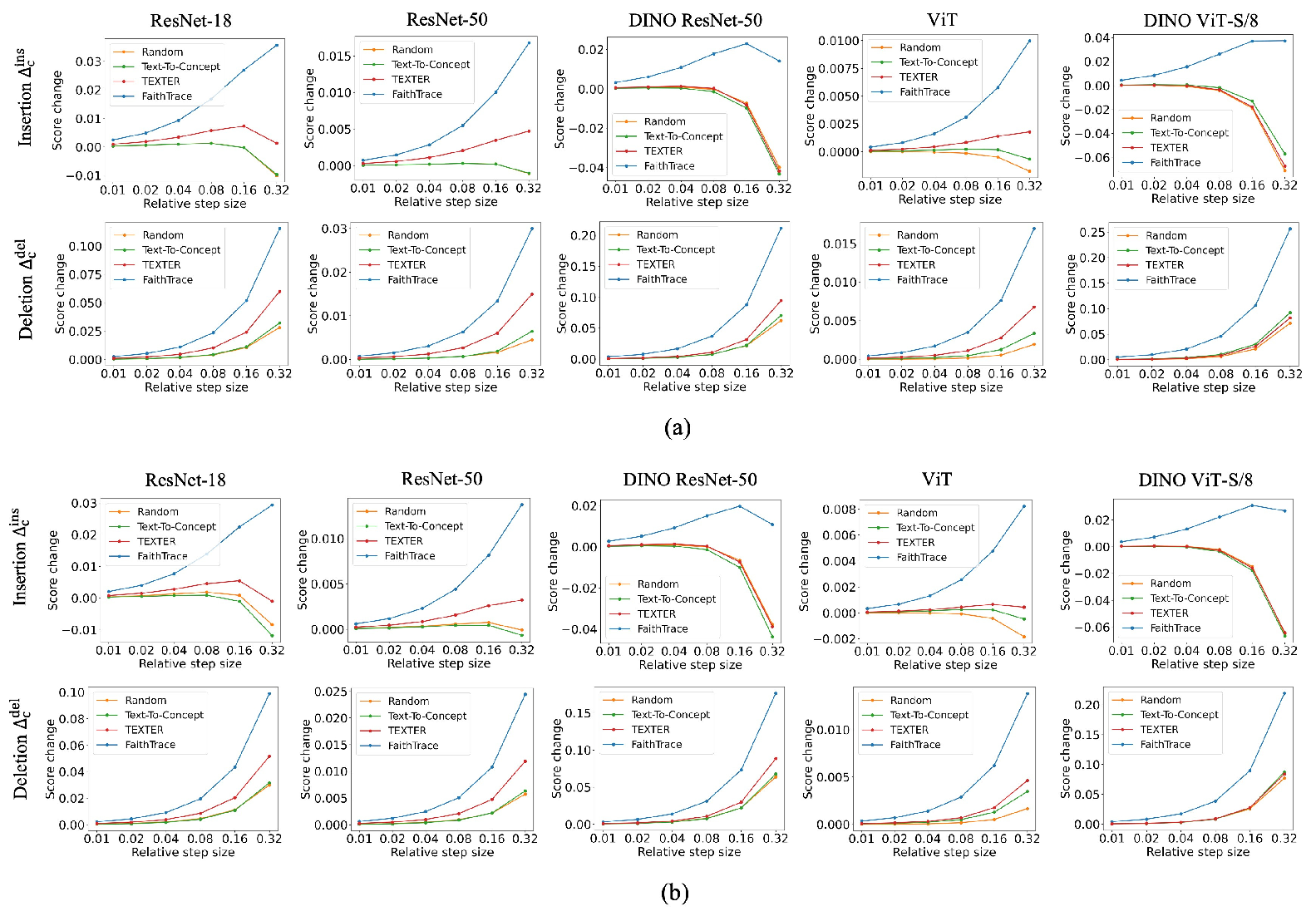}
  \caption{Influence curves for insertion and deletion under the top-1 (a) and top-5 (b) settings. At each step, larger positive values indicate higher faithfulness.}
  \label{fig:ins_del_more}
\end{figure*}

\begin{table*}[t]
\centering
\caption{
Sums over all steps in the influence curves for insertion (ins) and deletion (del)
under the top-1 and top-5 settings.
Higher is better. Best values are in bold.
All values are reported in units of $10^{-1}$ for readability.
}
\label{tab:ins_del_more}
\begin{tabular}{lllcccc}
\toprule
\textbf{Model} & \textbf{Top-$k$} & \textbf{Metric}
& {\small\textbf{Random}}
& {\small\textbf{Text-To-Concept}}
& {\small\textbf{TEXTER}}
& {\cellcolor{gray!12}\small\textbf{\mymethod}} \\
\midrule

\multirow{4}{*}{ResNet-18}
& \multirow{2}{*}{1} & Ins $\uparrow$ & -0.072 & -0.068 & 0.204 & \cellcolor{gray!12}\textbf{0.961} \\
&                       & Del $\uparrow$ & 0.454 & 0.507 & 1.021 & \cellcolor{gray!12}\textbf{2.096} \\
\cmidrule(lr){2-7}
& \multirow{2}{*}{5} & Ins $\uparrow$ & -0.034 & -0.105 & 0.140 & \cellcolor{gray!12}\textbf{0.793} \\
&                       & Del $\uparrow$ & 0.489 & 0.493 & 0.670 & \cellcolor{gray!12}\textbf{1.774} \\
\midrule

\multirow{4}{*}{ResNet-50}
& \multirow{2}{*}{1} & Ins $\uparrow$ & -0.002 & -0.001 & 0.122 & \cellcolor{gray!12}\textbf{0.373} \\
&                       & Del $\uparrow$ & 0.073 & 0.095 & 0.258 & \cellcolor{gray!12}\textbf{0.550} \\
\cmidrule(lr){2-7}
& \multirow{2}{*}{5} & Ins $\uparrow$ & 0.018 & 0.007 & 0.089 & \cellcolor{gray!12}\textbf{0.304} \\
&                       & Del $\uparrow$ & 0.095 & 0.099 & 0.203 & \cellcolor{gray!12}\textbf{0.445} \\
\midrule

\multirow{4}{*}{DINO ResNet-50}
& \multirow{2}{*}{1} & Ins $\uparrow$ & -0.454 & -0.534 & -0.469 & \cellcolor{gray!12}\textbf{0.750} \\
&                       & Del $\uparrow$ & 0.955 & 1.037 & 1.428 & \cellcolor{gray!12}\textbf{3.642} \\
\cmidrule(lr){2-7}
& \multirow{2}{*}{5} & Ins $\uparrow$ & -0.422 & -0.540 & -0.430 & \cellcolor{gray!12}\textbf{0.628} \\
&                       & Del $\uparrow$ & 0.978 & 1.009 & 1.349 & \cellcolor{gray!12}\textbf{3.036} \\
\midrule

\multirow{4}{*}{ViT}
& \multirow{2}{*}{1} & Ins $\uparrow$ & -0.025 & -21.047 & 0.047 & \cellcolor{gray!12}\textbf{0.217} \\
&                       & Del $\uparrow$ & 0.026 & 0.053 & 0.114 & \cellcolor{gray!12}\textbf{0.310} \\
\cmidrule(lr){2-7}
& \multirow{2}{*}{5} & Ins $\uparrow$ & -0.023 & 0.003 & 0.020 & \cellcolor{gray!12}\textbf{0.178} \\
&                       & Del $\uparrow$ & 0.024 & 0.056 & 0.076 & \cellcolor{gray!12}\textbf{0.253} \\
\midrule

\multirow{4}{*}{DINO ViT-S/8}
& \multirow{2}{*}{1} & Ins $\uparrow$ & -0.948 & -0.703 & -0.884 & \cellcolor{gray!12}\textbf{1.293} \\
&                       & Del $\uparrow$ & 1.009 & 1.376 & 1.193 & \cellcolor{gray!12}\textbf{4.433} \\
\cmidrule(lr){2-7}
& \multirow{2}{*}{5} & Ins $\uparrow$ & -0.811 & -0.881 & -0.821 & \cellcolor{gray!12}\textbf{1.036} \\
&                       & Del $\uparrow$ & 1.155 & 1.265 & 1.245 & \cellcolor{gray!12}\textbf{3.768} \\
\bottomrule
\end{tabular}
\end{table*}

\section{More quantitative results}
\label{sec:more_quantitative}

\subsection{Additional results of influence curve}
\label{sec:more_ins_del}
In \cref{sec:quantitative_evaluations}, 
we report the influence curve results under the top-3 setting. 
Here, we additionally show the results under the top-1 and top-5 settings.

\par
Figure~\ref{fig:ins_del_more} shows the insertion and deletion curves, 
where each point is averaged over all images and the retrieved explanations. 
At each step, larger positive insertion and deletion values indicate higher faithfulness, as they reflect stronger changes in favor of class $c$ along the explanation direction and its opposite. 
Similar to \cref{fig:ins_del}, across all step sizes, 
our method consistently exhibits larger changes than the baselines, 
indicating that it produces more faithful explanations.

\par
Table~\ref{tab:ins_del_more} shows the sums over all steps in the insertion and deletion curves,
where higher values are better.
Similar to \cref{tab:influence_curve},
the proposed method consistently outperforms the baselines by large margins,
indicating its effectiveness.

\begin{table*}[t]
\centering
\caption{Quantitative evaluation results for misclassification cases.
We report the directional score (Mean/NR) and the influence curve sums (Ins/Del, in units of $10^{-1}$) under the top-1 setting.
Best values are shown in bold.
}
\label{tab:miss_quantitative_top1}
\begin{tabular}{llcccc}
\toprule
\textbf{Model} & \textbf{Method} &
\multicolumn{2}{c}{\small\textbf{Directional score}} &
\multicolumn{2}{c}{\footnotesize\textbf{Influence curve}} \\
\cmidrule(lr){3-4}\cmidrule(lr){5-6}
& &
\textbf{Mean} $\uparrow$ & \textbf{NR} $\downarrow$ &
\textbf{Ins} $\uparrow$ & \textbf{Del} $\uparrow$ \\
\midrule
\multirow{4}{*}{ResNet-18}
& Random          & -0.041 & 0.640 &  0.171 & 1.663 \\
& Text-To-Concept &  0.053 & 0.300 & -0.489 & 1.215 \\
& TEXTER          &  0.075 & 0.225 &  0.450 & 2.485 \\
& \cellcolor{gray!12}\mymethod       & \cellcolor{gray!12}\textbf{0.227} & \cellcolor{gray!12}\textbf{0.000} & \cellcolor{gray!12}\textbf{2.725} & \cellcolor{gray!12}\textbf{5.400} \\
\midrule
\multirow{4}{*}{ResNet-50}
& Random          & -0.027 & 0.650 & 0.318 & 0.505 \\
& Text-To-Concept &  0.024 & 0.190 & 0.333 & 0.557 \\
& TEXTER          &  0.047 & 0.130 & 0.730 & 1.036 \\
& \cellcolor{gray!12}\mymethod       & \cellcolor{gray!12}\textbf{0.083} & \cellcolor{gray!12}\textbf{0.000} & \cellcolor{gray!12}\textbf{1.593} & \cellcolor{gray!12}\textbf{1.981} \\
\midrule
\multirow{4}{*}{DINO ResNet-50}
& Random          & -0.286 & 0.610 & 0.159 & 3.505 \\
& Text-To-Concept &  0.505 & 0.180 & -0.482 & 2.756 \\
& TEXTER          &  0.723 & 0.200 & -0.101 & 3.286 \\
& \cellcolor{gray!12}\mymethod       & \cellcolor{gray!12}\textbf{1.934} & \cellcolor{gray!12}\textbf{0.000} & \cellcolor{gray!12}\textbf{3.461} & \cellcolor{gray!12}\textbf{8.912} \\
\midrule
\multirow{4}{*}{ViT}
& Random          & -0.016 & 0.570 & 0.210 & 0.286 \\
& Text-To-Concept &  0.039 & 0.340 & 0.178 & 0.290 \\
& TEXTER          &  0.055 & 0.220 & 0.470 & 0.633 \\
& \cellcolor{gray!12}\mymethod       & \cellcolor{gray!12}\textbf{0.148} & \cellcolor{gray!12}\textbf{0.000} & \cellcolor{gray!12}\textbf{1.057} & \cellcolor{gray!12}\textbf{1.289} \\
\midrule
\multirow{4}{*}{DINO ViT-S/8}
& Random          & -0.022 & 0.740 & -1.853 & 2.382 \\
& Text-To-Concept &  0.003 & 0.490 & -1.995 & 3.051 \\
& TEXTER          & -0.012 & 0.680 & -1.850 & 2.309 \\
& \cellcolor{gray!12}\mymethod       & \cellcolor{gray!12}\textbf{0.052} & \cellcolor{gray!12}\textbf{0.010} & \cellcolor{gray!12}\textbf{3.986} & \cellcolor{gray!12}\textbf{8.808} \\
\bottomrule
\end{tabular}
\end{table*}

\begin{table*}[t]
\centering
\caption{Quantitative evaluation results for misclassification cases.
We report the directional score (Mean/NR) and the influence curve sums (Ins/Del, in units of $10^{-1}$) under the top-5 setting.
Best values are shown in bold.
}
\label{tab:miss_quantitative_top5}
\begin{tabular}{llcccc}
\toprule
\textbf{Model} & \textbf{Method} &
\multicolumn{2}{c}{\small\textbf{Directional score}} &
\multicolumn{2}{c}{\footnotesize\textbf{Influence curve}} \\
\cmidrule(lr){3-4}\cmidrule(lr){5-6}
& &
\textbf{Mean} $\uparrow$ & \textbf{NR} $\downarrow$ &
\textbf{Ins} $\uparrow$ & \textbf{Del} $\uparrow$ \\
\midrule
\multirow{4}{*}{ResNet-18}
& Random          & -0.042 & 0.680 &  0.135 & 1.518 \\
& Text-To-Concept &  0.051 & 0.314 & -0.184 & 1.552 \\
& TEXTER          &  0.063 & 0.284 &  0.652 & 2.650 \\
& \cellcolor{gray!12}\mymethod       & \cellcolor{gray!12}\textbf{0.171} & \cellcolor{gray!12}\textbf{0.000} & \cellcolor{gray!12}\textbf{2.271} & \cellcolor{gray!12}\textbf{4.598} \\
\midrule
\multirow{4}{*}{ResNet-50}
& Random          & -0.026 & 0.712 & 0.339 & 0.510 \\
& Text-To-Concept &  0.019 & 0.286 & 0.229 & 0.436 \\
& TEXTER          &  0.030 & 0.230 & 0.598 & 0.853 \\
& \cellcolor{gray!12}\mymethod       & \cellcolor{gray!12}\textbf{0.062} & \cellcolor{gray!12}\textbf{0.014} & \cellcolor{gray!12}\textbf{1.319} & \cellcolor{gray!12}\textbf{1.656} \\
\midrule
\multirow{4}{*}{DINO ResNet-50}
& Random          & -0.212 & 0.604 & -0.109 & 3.319 \\
& Text-To-Concept &  0.507 & 0.248 & -0.281 & 3.189 \\
& TEXTER          &  0.643 & 0.232 & -0.141 & 3.464 \\
& \cellcolor{gray!12}\mymethod       & \cellcolor{gray!12}\textbf{1.574} & \cellcolor{gray!12}\textbf{0.002} & \cellcolor{gray!12}\textbf{2.714} & \cellcolor{gray!12}\textbf{7.603} \\
\midrule
\multirow{4}{*}{ViT}
& Random          & -0.022 & 0.638 & 0.161 & 0.276 \\
& Text-To-Concept &  0.039 & 0.296 & 0.165 & 0.285 \\
& TEXTER          &  0.038 & 0.300 & 0.328 & 0.465 \\
& \cellcolor{gray!12}\mymethod       & \cellcolor{gray!12}\textbf{0.121} & \cellcolor{gray!12}\textbf{0.000} & \cellcolor{gray!12}\textbf{0.779} & \cellcolor{gray!12}\textbf{0.967} \\
\midrule
\multirow{4}{*}{DINO ViT-S/8}
& Random          & -0.024 & 0.740 & -2.156 & 2.237 \\
& Text-To-Concept &  -0.000 & 0.524 & -2.127 & 2.770 \\
& TEXTER          & -0.011 & 0.630 & -2.072 & 2.330 \\
& \cellcolor{gray!12}\mymethod       & \cellcolor{gray!12}\textbf{0.039} & \cellcolor{gray!12}\textbf{0.048} & \cellcolor{gray!12}\textbf{2.806} & \cellcolor{gray!12}\textbf{7.435} \\
\bottomrule
\end{tabular}
\end{table*}

\subsection{Additional results on misclassification cases}
\label{sec:more_miss}
We report quantitative results for misclassification cases under the top-3 setting in the main paper.
Here, we additionally report results under the top-1 and top-5 settings.
Following the main paper, we evaluate faithfulness using the directional score (Mean/NR) and the influence curve (insertion/deletion sums) on 100 misclassified images per model.

\par
Tables~\ref{tab:miss_quantitative_top1} and~\ref{tab:miss_quantitative_top5} summarize the results.
Across all evaluated models, \mymethod achieves the best performance under both settings, consistent with the top-3 results in \cref{tab:miss_quantitative}.
These results further support that \mymethod provides explanations that better align with the classifier's decision behavior across different top-$k$ configurations.

\end{document}